\documentclass{article}

\usepackage{microtype}
\usepackage{subcaption}
\usepackage{graphicx}
\usepackage{booktabs} 
\usepackage{multirow}
\usepackage{threeparttable}
\usepackage[inline]{enumitem}
\usepackage{balance}
\usepackage[utf8]{inputenc}
\usepackage{xspace}
\usepackage[numbers]{natbib}
\usepackage[a4paper,margin=1.1in,footskip=0.25in]{geometry}
\usepackage[colorlinks]{hyperref}
\usepackage[export]{adjustbox}
\usepackage{subcaption}

\makeatletter
\renewcommand{\paragraph}{%
  \@startsection{paragraph}{4}%
  {\z@}{0.5ex \@plus 1ex \@minus .2ex}{-1em}%
  {\normalfont\normalsize\bfseries}%
}
\makeatother

\usepackage[table,xcdraw]{xcolor}

\usepackage{amsmath}
\usepackage{amssymb}
\usepackage{mathtools}
\usepackage{amsthm}
\usepackage[capitalize,noabbrev]{cleveref}

\theoremstyle{plain}

\theoremstyle{definition}

\theoremstyle{remark}

\renewcommand{\cite}{\citep}
\hypersetup{citecolor=green}

\usepackage[final]{changes}
\renewcommand{\added}[1]{#1}

\title{Evaluating the Robustness of the ``Ensemble Everything Everywhere'' Defense}
\date{}
\author{
 Jie Zhang$^1$  \quad Christian Schlarmann$^2$ \quad Kristina Nikolić$^1$ \\ \quad Nicholas Carlini$^3$  \quad Francesco Croce$^4$ \quad Matthias Hein$^2$ \quad Florian Tramèr$^1$ \vspace{1.0em}\\
  \small $^1${ETH Zurich} \hspace{1.0em} 
  $^2${University of Tübingen}\hspace{1.0em} $^3${Google DeepMind} \hspace{1.0em} $^4${EPFL}
}

\begin{document}

\maketitle

\textcolor{red}{An earlier version of this report by Zhang, Nikolić, Carlini and Tramèr contained incorrect attack results due to a bug that created perturbations $10\times$ larger than the bound allowed by the defense~\cite{fort2025noteimplementationerrorsrecent}. This new version corrects this bug and the resulting claims about the defense's robustness. We also add results from a concurrent robustness evaluation by Schlarmann, Croce and Hein~\cite{heincomment}.}

\begin{abstract}
\emph{Ensemble everything everywhere} is a defense to adversarial examples that was recently proposed to make image classifiers robust.
This defense works by ensembling a model's intermediate representations at multiple noisy image resolutions, producing a single robust classification.
This defense was shown to be effective against multiple state-of-the-art attacks. Perhaps even more convincingly,
it was shown that the model's gradients are \emph{perceptually aligned}: attacks against the model
produce noise that perceptually resembles the targeted class.

In this short note, we show that this defense' robustness to adversarial attack \added{is overclaimed}.
We first show that the defense's randomness and ensembling method cause severe gradient masking.
We then use standard adaptive attack techniques to reduce the robust accuracy of the defense from $48\%$ to 
\added{$14\%$}
on CIFAR-100 and from $62\%$ to 
\added{$11\%$}
on CIFAR-10, under the $\ell_\infty$ norm threat model with $\varepsilon=8/255$. 

\end{abstract}

\section{Introduction}

\emph{Ensemble everything everywhere} \cite{fort2024ensemble} is a recent defense to adversarial examples \cite{biggio2013evasion, szegedy2013intriguing}.
Given any input image, this model applies random transformations to the input at different resolutions and then ensembles the inner activation layers to produce a final robust prediction;
by doing this, the authors claim a robust accuracy of $72\%$ on CIFAR-10, and $48\%$ on CIFAR-100 under standard threat models (the authors show that the robustness can be further boosted using adversarial training, to achieve state-of-the-art robustness).

This is certainly not the first proposed defense that claims high (or even state-of-the-art) robustness by combining similar techniques.
And such empirical defenses---with the notably exception of those based on adversarial training~\cite{madry2018towards}---all end up broken without exception.

So why should we hope that this defense will be different?
Perhaps most convincingly, the defense is shown to produce \emph{perceptually aligned} perturbations. That is, when starting from a blank image and running an unbounded attack that optimizes for a target class (say ``cat''), the attack produces an image that has cat-like features.
This suggests two things: (1) the attack can successfully optimize over the defense (at least in the large perturbation regime), suggesting that gradient masking might not be a concern~\cite{athalye2018obfuscated}; (2) the defense seems to rely on perceptually aligned features, a property that had previously been observed in robust models~\cite{tsipras2018robustness}.
In addition, the defense is evaluated against a state-of-the-art attack (AutoAttack~\cite{croce2020reliable}) 
and achieves high (but not implausible) robustness.

Unfortunately, we show that this defense is not \added{as} robust \added{as initially claimed} and that the existing evaluation does suffer from gradient masking.
We build a stronger \emph{adaptive attack} that reduces the robust accuracy to \added{$14\%$} for CIFAR-100, or 
\added{$11\%$} for CIFAR-10.%

In summary, while the approach proposed in~\cite{fort2024ensemble} is an interesting and novel way to yield informative visualizations and image manipulations, it does not yield a defense that \added{achieves high robustness}  
to worst case perturbations.

\added{These results are reminiscent of a previous defense---ME-Net~\cite{yang2019me}---which also yields interpretable perturbations but was shown to have only low robustness on CIFAR-10 under adaptive attacks~\cite{tramer2020adaptive}.}

\section{Background}

A defense to adversarial examples~\cite{szegedy2013intriguing, biggio2013evasion} is a classifier $f$ designed so that,
for any ``small'' perturbation $\delta$ to an evaluation example $(x, y)$, the model assigns the correct label $f(x + \delta) = y$.
For the purpose of this paper, we focus on $\ell_\infty$ bounded attacks
which require that $\lVert \delta \rVert_\infty < \varepsilon$.

There is an exceptionally long line of work studying robustness to exactly
these types of attacks.
But almost all of these papers are later shown to be broken by stronger
attacks (see, e.g.,~\cite{carlini2017towards, athalye2018obfuscated, tramer2020adaptive}).

 (1)
adversarial training~\cite{szegedy2013intriguing, madry2018towards}, which trains a model on adversarial examples until the model is (empirically) robust to attacks;
(2) robustness certification~\cite{raghunathan2018certified, wong2018provable, gehr2018ai}, which computes and optimizes a tractable lower bound on a model's robustness to obtain a provable guarantee; (3) randomized smoothing~\cite{cohen2019certified, lecuyer2019certified}, which trains a model on noisy images
and then ensembles many (e.g., 1000) noisy predictions at test time to provide a probabilistic robustness guarantee.

If effective, \emph{ensemble everything everywhere} would represent a new and fourth approach to achieving robustness.
Although this defense idea is not completely original (many prior defenses have claimed robustness
by studying hidden layer activations~\cite{metzen2017detecting},
ensembling many predictors~\cite{pang2019improving, abbasi2017robustness, xu2017feature},
adding randomness to the input image~\cite{xie2017mitigating},
or applying other input transformations \cite{raff2019barrage, guo2017countering})
this defense has four compelling properties:
\begin{enumerate}
    \item \emph{High clean accuracy:} $>90\%$ test accuracy in CIFAR-10, higher than most other adversarially robust defenses.
    \item \emph{High (claimed) robust accuracy:} close or exceeding SOTA on CIFAR-10 and CIFAR-100 for $\ell_\infty$ attacks bounded by $\epsilon=8/255$.
    \item \emph{Efficient inference:} The defense uses a single forward pass through a standard convolutional model (it is therefore orders of magnitude cheaper than random smoothing).
    \item \emph{Efficient training:} The defense requires no adversarial training.
\end{enumerate}

\subsection{Ensemble Everything Everywhere}

This defense combines three ideas:

\begin{enumerate}
    \item \emph{Multi-resolution:} Instead of processing a single image $x \in [0,1]^{d \times d}$, the defense first stacks multiple copies of the input at different resolutions. 
    
    Given a set of resolutions $d_1, \dots, d_k \leq d$ we define
    \[
        x_k = \textsc{upscale}\Big(\textsc{downscale}(x, d_i) + \mathcal{N}(0, \sigma_1^2 \cdot I_{d_i \times d_i})\Big) + \mathcal{N}(0, \sigma_2^2 \cdot I_{d \times d}) \;.
    \]
    That is, the input is downscaled to size $d_i \times d_i$, then noised, then upscaled to size $d \times d$, and noised again.
    The authors' implementation adds other random transforms (e.g., jitter, contrast, color shifts) to the input, which we omit here for simplicity.
    The input to the model is $x_{\text{multi}} = [x_1, x_2, \dots, x_k]$.

    \item \emph{Ensembling intermediate layers:} The defense also ensembles features at multiple intermediate layers.
    
    Denote an $n$-layer model as $f(x) = g_n \circ g_{n-1} \circ \dots g_2 \circ g_1(x)$, where $g_i$ is the $i$-th layer. Define $f_i$ as the partial function that runs the model up to layer $i$.
    The defense jointly trains multiple classifiers of the form $h_i \circ f_i(x)$, where $h_i$ is a linear layer that maps the features at layer $i$ to a vector of logits.
    At inference time, the defense aggregates the logits for a set of layers $i \in I$:
    \[
        f_{\text{ens}}(x_{\text{multi}}) = \textsc{aggregate}\big(\{h_i \circ f_i (x_{\text{multi}}) \}_{i \in I)}\big)
    \]
    A simple aggregation function is to just take the mean of all the intermediate logits.

    \item \emph{Robust aggregation:}
    Finally, the defense proposes a more robust aggregation function for intermediate logits. Given the set of logits $Z \in \mathbb{R}^{|I| \times C}$ (where $|I|$ is the number of ensembled layers, and $C$ is the number of classes),
    the \textsc{CrossMax} aggregation function is defined as:
    \begin{align*}
        \textsc{CrossMax}(Z):\quad & Z \gets  Z - \max(Z, \mathtt{axis}=1)\\
            &Z \gets  Z - \max(Z, \mathtt{axis}=0)\\
            &Z \gets \mathtt{sort}(Z, \mathtt{axis}=0)\\
            &\mathbf{output}\ Z[k] 
    \end{align*}
    In words, we first normalize all intermediate predictions so that the most likely class predicted by each layer has score 0 (i.e., the largest value in each \emph{row} is equal to 0).
    Then, we normalize class predictions by subtracting the most confident prediction for each class across all layers (i.e., the largest value in each \emph{column} is equal to 0).
    Finally, for each class, we return the \emph{k\textsuperscript{th}-highest} normalized score for that class across all layers ($k=3$ in the released code).
\end{enumerate}

\section{Existing Tests Reveal Gradient Masking} \label{s:gradient_masking}

\emph{Gradient Masking} \cite{papernot2016practical, tramer2018ensemble, athalye2018obfuscated} is one of the most common
reasons why an adversarial example defense appears effective but ends up being broken.
A defense is said to mask its gradients if the local gradient direction does not usefully align with the direction of the nearest adversarial examples.
For example, by saturating the output softmax of a classifier, a defense could cause numerical instabilities that lead to zero loss and gradients~\cite{carlini2016defensive}.
Since gradient masking is such a prevalent issue, there are now numerous tests designed to identify it.
We find that two of these tests would have given strong evidence that this defense's original evaluation was untrustworthy because of significant gradient masking.

\subsection{Visualizing the Loss Landscape}

We begin with a simple method to understand whether or not the defense's loss landscape is amenable to gradient-based optimization.
Given a point $x$, we pick two directions $d_1, d_2$ in the input space (where one direction points towards an adversarial example, and the other is randomly chosen to be orthogonal to the first one), and measure the model's loss in a two-dimensional slice of the form $x + \alpha \cdot d_1 + \beta \cdot d_2$.

\begin{figure}[t]
    \centering
    \begin{subfigure}[b]{0.32\textwidth}
        \centering
        \includegraphics[width=\textwidth,trim={.1in 1in 2in 1.4in}, clip]{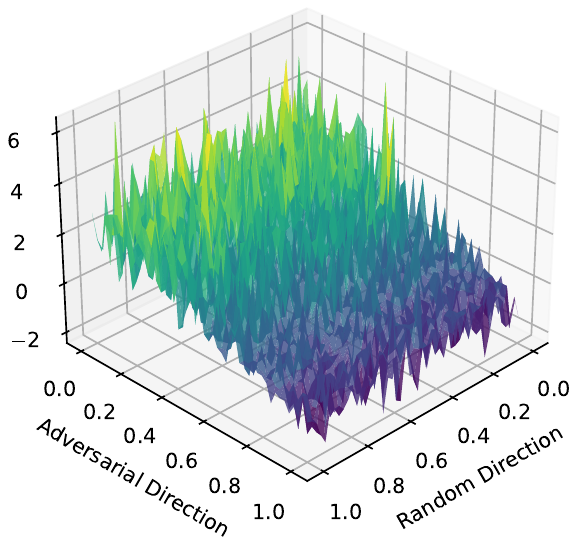}
        \caption{Loss of original model}
        \label{fig:loss_noisy}
    \end{subfigure}
    \hfill
    \begin{subfigure}[b]{0.32\textwidth}
        \centering
        \includegraphics[width=\textwidth,trim={.1in 1in 2in 1.4in}, clip]{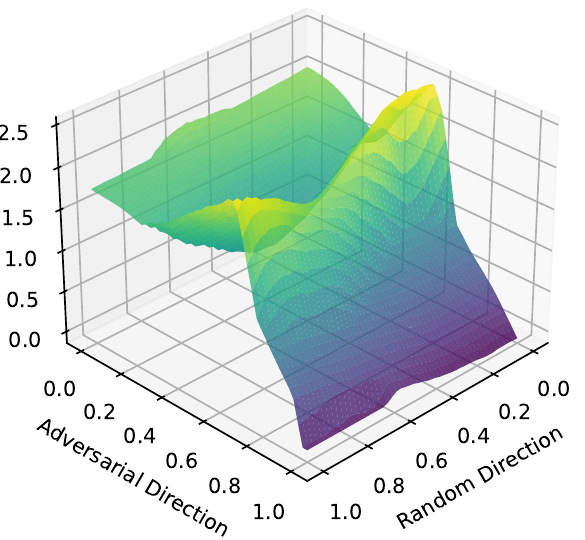}
        \caption{Loss of model with randomness disabled}
        \label{fig:loss_clean}
    \end{subfigure}
    \hfill
    \begin{subfigure}[b]{0.32\textwidth}
        \centering
        \includegraphics[width=\textwidth,trim={.1in 1in 2in 1.4in}, clip]{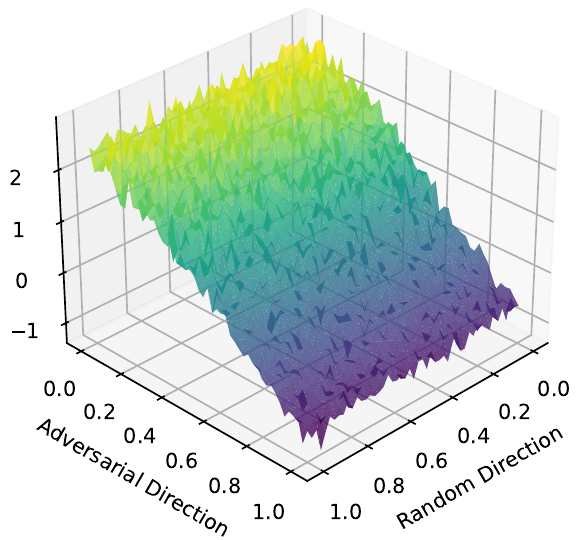}
        \caption{Loss of model, averaged over many evaluations}
        \label{fig:loss_avg}
    \end{subfigure}
    \caption{If we plot a two-dimensional slice of the loss surface, the original model in (a) has extremely large spikes in the logprobs of the model output. 
    This makes it very difficult for gradient-based search to identify adversarial examples.
    If we remove randomness, in (b) we see that the loss surface is indeed smooth, confirming that the model itself does not introduce gradient instabilities.
    Therefore, by ensembling over randomness in (c) we can correct for the noisy loss surface and break the defense.}
    \label{fig:loss_comparison}
\end{figure}

The loss surface plotted in Figure~\ref{fig:loss_noisy} is extremely ``spiky'' with a large number of local minima that hinder gradient descent.
To understand the cause of this effect, we first disable the model's randomness (by picking the same random pre-processing for each inference). This leads to a much smoother loss surface (Figure~\ref{fig:loss_clean}), but still with a significant local minimum.
Finally, we average out the randomness using an Expectation-over-transformation approach~\cite{athalye2018obfuscated}, by averaging the loss over multiple inferences (Figure~\ref{fig:loss_avg}). This partially smooths out the loss, and reveals a more stable optimization landscape on a macro-scale, but with small spikes remaining on a local scale.

Overall, this simple visualization shows that we have little hope of finding adversarial examples if we apply naive gradient descent directly to the full defense.

\subsection{Unit Tests}

Zimmerman \emph{et al.} propose a simple method to determine if an evaluation of
a defense to adversarial examples is performed correctly~\cite{zimmermann2022increasing}.
Given the trained classifier $f$, begin by re-initializing the final layer of the
model so that the output predictions are completely arbitrary and have no inherent meaning. Then, re-train the final layer so that for every input $x$ there is \emph{always} another nearby ``adversarial'' example $x'$ that has a different classification.\footnote{We omit a number of details. 
In practice we condense the model to have two output classes, and always attack from class 0 to class 1.
The bias should be set so that random search does not trivially find adversarial examples.}
Now, use the attack from the original evaluation to find adversarial examples on this classifier.
If the attack fails to find an adversarial example for some point $x$, then we know it must be too weak to be correct, because we have constructed the model to have this property.

Performing this test with the original evaluation code results in a unit test pass rate of just $60\%$.
This suggests that the original evaluation was insufficiently strong to
correctly evaluate whether or not the defense was effective.
Thus, the remainder of this paper will be dedicated to developing a stronger
attack that can correctly account for the randomness inherent in this defense.

\section{Evaluating the Defense's Robustness} 
\label{s:breaking_defence}

\paragraph{Setup.} We use the official codebase released by the authors\footnote{\url{https://github.com/stanislavfort/ensemble-everything-everywhere}} to train the defense on CIFAR-10 and CIFAR-100.
Our models achieve clean accuracy of $88.9\%$ on CIFAR-10, and $64.1\%$ on CIFAR-100. We then run the AutoAttack evaluation considered by the authors, and obtain robust accuracies of $61.8\%$ on CIFAR-10 and $47.9\%$ on CIFAR-100 (see Table~\ref{table:results}). This is consistent with the results reported by the authors in their repository.\footnote{The paper~\cite{fort2024ensemble} reports slightly higher robustness numbers using larger models, longer training runs, and combining with adversarial training. As these models and their training code are not publicly released, we rely on the official training code.} Since this defense relies heavily on randomness, we conduct 10 runs for all results and then report the mean and standard deviation.
\added{In contrast to~\cite{fort2024ensemble}, we attack \textit{all} samples, even if they are classified incorrectly already in clean evaluation. Since the forward pass of the target model is subject to randomness, such samples are not necessarily \textit{always} classified incorrectly. Thus it is important to attack also these samples~\cite{liu2024towards}.}

\paragraph{Our \added{PGD} attack.} 
Our attack is rather simple: the main trick is to attack the defense without the \textsc{CrossMax} operator, which causes gradient masking.
We summarize our attack steps below:

\begin{itemize}
    \item \emph{Standard PGD:} We start with a standard PGD attack~\cite{madry2018towards} with 500 steps. Surprisingly, this already improves upon the results from AutoAttack. The reason is likely because we use a large number of steps which helps deal with the large randomness in the defense's pre-processing.
    \item \emph{Transfer from a model without \textsc{CrossMax}:} we found that if we replace the \textsc{CrossMax} aggregation by a simple mean, we can reliably reduce the model's robust accuracy. This suggests that either (1) all the robustness comes from the \textsc{CrossMax}, or (2) the \textsc{CrossMax} causes gradient masking. We find that the latter is true: if we take the adversarial examples optimized for the model with a mean aggregation, and then simply transfer these to the target model, the robust accuracy is reduced by about $1/3$.\footnote{This was also described in~\cite{heincomment}.}
    \item \emph{Expectation over Transformation (EoT):} Since the defense uses a lot of randomness, we apply the standard EoT trick~\cite{athalye2018obfuscated} where we approximate the expected value of the gradient by performing multiple backward passes with different randomness.
    
    \item \added{\emph{Bag of Tricks}: A standard PGD+EoT attack transferred from a model without \textsc{CrossMax} works reasonably well, and reduces robustness to around 20--30\%. We then add further simple tricks---additional PGD steps (up to 400), more EoT steps (up to 100), a change of loss function (the Hinge loss of~\cite{carlini2017towards} instead of a standard cross-entropy)---to reduce the robustness further to around $11.3\%$ on CIFAR-10 and $13.8\%$ on CIFAR-100.}

\end{itemize}

\begin{table}[h]
\centering
\caption{We report the robust accuracy on 100 test samples under an \( \ell_\infty = 8/255 \) attack strength. This table presents the results of attacks on multi-resolution ResNet152 models fine-tuned on CIFAR-10 and CIFAR-100 from an ImageNet-pretrained model, without any adversarial training.}
\label{table:results}
\vspace{5pt}
\begin{tabular}{@{}l r r@{}}
\toprule
& \multicolumn{2}{c}{Accuracy (\%)}\\
\cmidrule{2-3}
Attack & CIFAR-10 & CIFAR-100\\
\midrule
None    & $88.9 \pm 2.8$ & $64.1 \pm 2.4$\\
AutoAttack & $61.8 \pm 2.3$ & $47.9 \pm 2.7$  \\ 
\midrule
PGD        & $54.0 \pm 2.0$ & $34.6 \pm 4.0$     \\
+ transfer & $32.6 \pm 1.9$ & $22.2 \pm 2.1$     \\
+ EoT      & $27.5 \pm 2.3$ & $19.5 \pm 1.5$     \\
\added{+ bag of tricks} &  \added{$11.3 \pm 2.5$} & \added{$13.8 \pm 2.1$} \\
\midrule
\added{APGD w/ transfer, EoT} & \added{$16.4 \pm 2.2$} & \added{$16.8 \pm 1.3$} \\
\added{+ larger initial radius} & \added{$15.3 \pm 2.7$} & \added{$16.0 \pm 1.5$} \\
\added{+ 1000 total steps} & \added{$13.8 \pm 2.5$} & \added{$15.9 \pm 1.7$} \\
\midrule
\added{\textbf{Best-of-both}} & \added{$\mathbf{10.9 \pm 2.4}$} & \added{$\mathbf{13.6 \pm 1.4}$} \\
\bottomrule
\end{tabular}
\end{table}

\begin{figure}[h]
    \centering
    \begin{subfigure}{0.5\linewidth}
        \includegraphics[width=\linewidth]{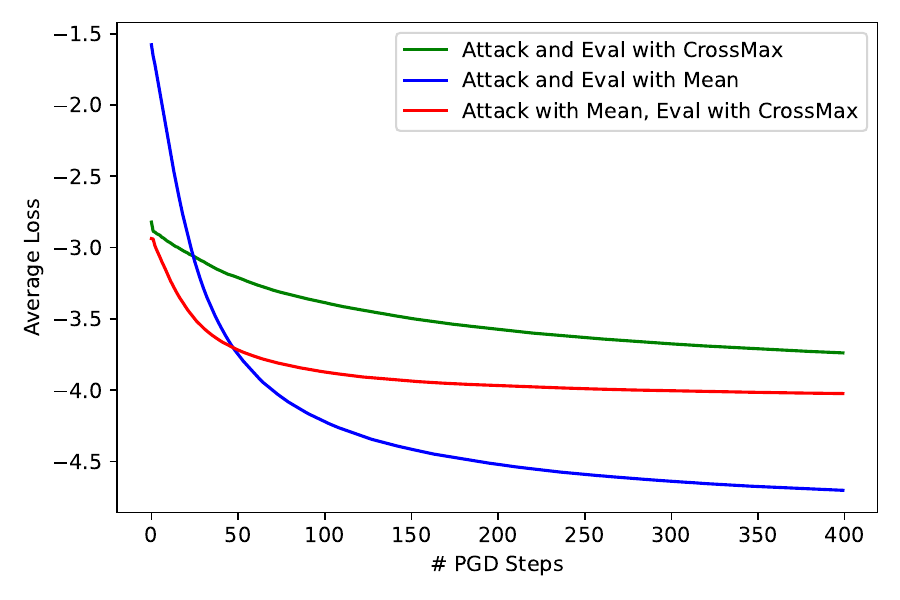}
        \caption{}
    \end{subfigure}%
    \begin{subfigure}{0.5\linewidth}
        \includegraphics[width=\linewidth]{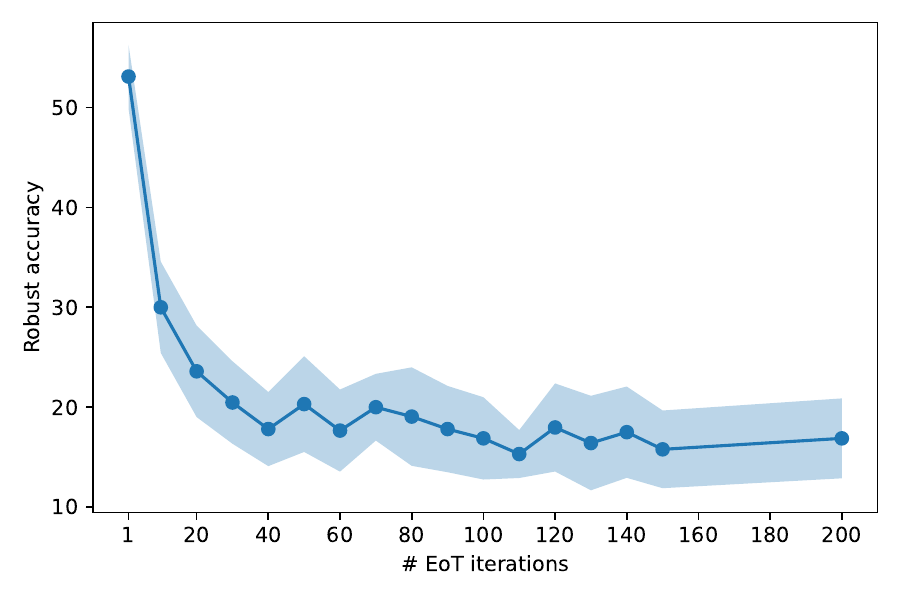}
        \caption{}
    \end{subfigure}
    \caption{\textbf{(a)} For different attack strategies, we plot the attack's loss averaged over 100 examples: (1) directly optimizing the target model (in green) has trouble converging due to gradient masking; (2) attacking the ``source'' model with a mean aggregation (blue) works well; (3) transferring the attack from the source model to the target model (red) outperforms the attack that directly optimizes over the target model.
    \added{\textbf{(b)} We show an ablation on the amount of EoT iterations with the APGD attack on 32 CIFAR-10 samples using 100 steps. We observe that it is crucial to use a sufficiently high amount of iterations, however, there is no consistent improvement beyond 100 EoT iterations.}
    }
    \label{fig:attack}
\end{figure}

We visualize the performance of the different attack steps in Figure~\ref{fig:attack}.

\paragraph{Our APGD attack.} 
\added{This attack is based on APGD~\cite{croce2020reliable}, which is known to be a strong attack due to its adaptive step size schedule. We maximize the cross-entropy loss, using 100 steps by default. To adapt this attack to the target model at hand, we use the following approach:}
\begin{itemize}
    \item \added{\emph{Transfer from a model without \textsc{CrossMax}:} Similar to our PGD attack, we utilize transfer from a model that replaces \textsc{CrossMax} with mean aggregation.}
    
    \item \added{\emph{Expectation over Transformation (EoT):} We use 100 EoT iterations. An ablation on the amount of EoT iterations is shown in Figure~\ref{fig:attack}.}
    
    \item \added{\emph{Larger initial radius:} We use a larger perturbation budget at the beginning of the attack as proposed in~\cite{croce2021mind}. This allows the attack to find promising adversarial directions more easily. The perturbation radius is then gradually decreased during the attack according to the following schedule: $3\epsilon$ for $30\%$ of total iterations, $2\epsilon$ for $30\%$ of total iterations, and $\epsilon$ for $40\%$ of total iterations. This is an existing functionality of APGD that is by default deactivated.}
\end{itemize}

\section{Discussion}

Unfortunately, this defense did not end up giving us a fourth category of approaches for \added{(high)} adversarial robustness.
We believe there are a few lessons to be learned from this defense and the (cautious) enthusiasm that it originally caused.

\begin{figure}[t]
    \centering
    \begin{subfigure}[t]{0.95\textwidth}
        \centering
        \includegraphics[width=\textwidth]{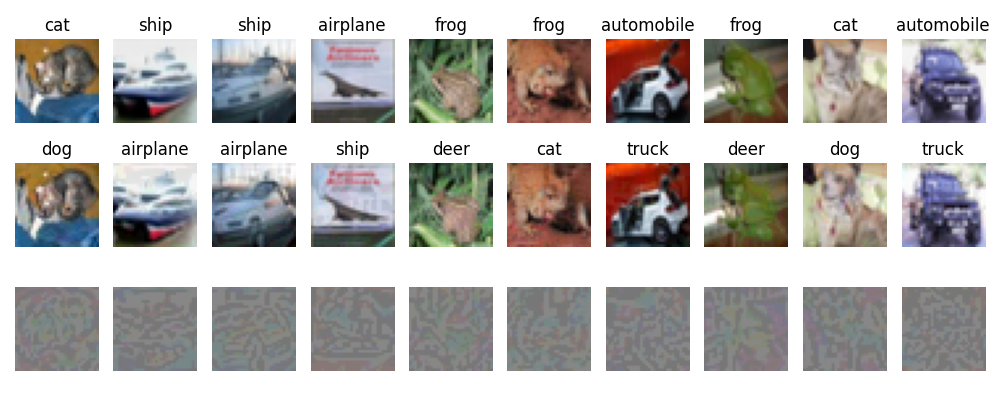}
        \vspace{-2em}
        \caption{CIFAR-10}
    \end{subfigure}%
    \\[1.5em]
    \begin{subfigure}[t]{0.95\textwidth}
        \centering
        \includegraphics[width=\textwidth]{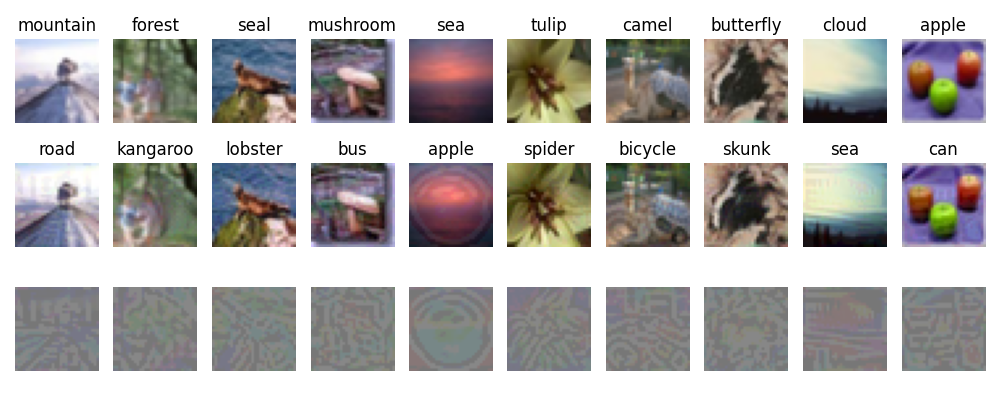}
        \vspace{-2em}
        \caption{CIFAR-100}
    \end{subfigure}
    \caption{\added{Examples of clean images (top row) with the adversarial examples found by our attack (middle row), and the corresponding perturbation (bottom row, centered and magnified). In some cases, the perturbations are clearly interpretable, e.g, CIFAR-10 ``cat'' $\to$ ``dog'' (2nd from right) or CIFAR-100 ``sea'' $\to$ ``apple'' (5th from left).}}
    \label{fig:examples}
\end{figure}

\begin{itemize}
    \item This defense mostly combines and extends past ideas that did not work: aggregating multiple viewpoints~\cite{luo2015foveation}, random input transformations~\cite{guo2017countering}, ``biologically-inspired'' visual processing~\cite{nayebi2017biologically, brendel2017comment}, aggregation of internal features~\cite{sabour2015adversarial}, etc.

    But this does not mean that revisiting past failed ideas is a bad thing. For example, both standard adversarial training~\cite{madry2018towards} and fast adversarial training~\cite{wong2020fast} carefully revisited ideas that had previously been attempted and incorrectly dismissed.
    
    However, given such past failures, new empirical defenses need a high degree of evidence to convince the community.

    \item We could analyze and \added{partially} break this defense using only existing techniques~\cite{zimmermann2022increasing, tramer2020adaptive, athalye2018obfuscated}.
    As already noted a few years ago in \cite{tramer2020adaptive}, the time seems to have passed where breaking a defense requires new research ideas. Now, it is mainly an exercise in rigorously applying known techniques until the defense breaks (or one fails after exhausting all attempts). 
    The issue with current defense evaluations is thus primarily \emph{methodological}: defense authors should try as hard as possible to cover all possible attack strategies before claiming empirical robustness.
    
    \item Although our attack did not need any new techniques, it does highlight (again) the importance of performing careful \emph{adaptive} evaluations.
    Although it is tempting to run standard off-the-shelf evaluation tools such as AutoAttack~\cite{croce2020reliable}, this is not sufficient.
    As we show above, AutoAttack maxes out at only a $52\%$ attack success rate on CIFAR-100, while our adaptive attack achieves $\geq 85\%$ success.

    \item Identifying these types of flaws requires the use of specialized tools designed specifically for this purpose.
    The original paper attempts to argue robustness, in part, by showing that
    the defended models have interpretable gradients.
    This is because many robust models indeed have interpretable gradients~\cite{tsipras2018robustness}.
    But the converse is not necessarily true~\cite{ganz2022perceptually}---as we
    showed again in the case of this model.
    Its gradients are highly perceptually aligned,
    but the model exhibits only very small robustness to adversarial attacks.
    \added{A similar phenomenon had been observed with a prior defense---ME-Net~\cite{yang2019me}---which uses matrix estimation techniques to yield interpretable gradients, but only very moderate adversarial robustness~\cite{tramer2020adaptive}.}

    \item \added{Attacks have to be rigorously tested too, and adaptive evaluations can contain their own bugs!
    The implementation in the original version of this report  stacked multiple attack stages, and incorrectly computed the $\ell_\infty$ bound for each stage with respect to the output of the previous stage. As a result, after $k$ stages the adversarial perturbations could be as large as $k\cdot 8/255$. (see~\cite{fort2025noteimplementationerrorsrecent} for details.}
    
    \added{A good practice to avoid such errors (and which we used in our amended PGD attack) was to implement the attack in a script that saves the final adversarial examples to disk, and then have a second standalone script 
    that solely evaluates the attack against the original defense and data.}
    
    \added{We also used this script to visualize adversarial examples (see Figure~\ref{fig:examples}). While some of these showcase interpretable perturbations (as claimed in the original defense paper), others do not seem to.}
\end{itemize}

On the whole, our work re-iterates the need for very careful evaluations of defenses against adversarial examples.
While heuristic arguments based on the human visual system or on model interpretability can be useful to build intuition for a defense, they are by no means a trustworthy signal of robustness.

That being said, the ideas behind this defense are by far not without merit. The multi-resolution prior proposed by Fort and Lakshminarayanan~\cite{fort2024ensemble} appears to have many useful applications beyond worst-case adversarial robustness, such as for model interpretability and visualization, simple and highly controllable image generation and manipulation from a single CLIP model, and even for creating transferable adversarial examples for state-of-the-art multimodal LLMs like GPT-4~\cite{fort2024ensemble}.

\paragraph{\added{Acknowledgments.}}
\added{We thank Stanislav Fort for discovering the bug in the original implementation of the attack of Zhang et al., and for valuable discussions about the defense and comments about this write-up.}

\bibliography{references.bib}
\bibliographystyle{plain}

\end{document}